\documentclass[10pt,journal,compsoc]{./IEEEtran}

\usepackage{amsmath}
\usepackage{amstext}
\usepackage{amsfonts}
\usepackage{algorithm}
\usepackage{algpseudocode}
\usepackage[pdftex]{graphicx}
\usepackage{booktabs}
\usepackage{multirow}
\usepackage[tight,footnotesize]{subfigure}
\usepackage{url}
\usepackage{lastpage}
\usepackage{flushend}

\ifCLASSOPTIONcompsoc
  \usepackage[nocompress]{cite}
\else
  \usepackage{cite}
\fi

\newcommand\E{{\cal E}}
\newcommand\D{{\cal D}}
\newcommand\Ob{{\cal O}}
\newcommand\N{{\mathbb N}}
\newcommand\R{{\mathbb R}}

\newcommand\vect[1]{\mathbf{#1}}

\newcommand\A{{\vect A}}
\newcommand\W{{\vect W}}

\newcommand\test{^{\text{test}}}
\newcommand\train{^{\text{train}}}
\newcommand\val{^{\text{val}}}
\DeclareMathOperator*{\argmin}{arg\,min}

\begin{document}

\title{Multi-Relational Learning at Scale with ADMM 
}


\author{Lucas~Drumond,
        Ernesto~Diaz-Aviles,
        and~Lars~Schmidt-Thieme,
\IEEEcompsocitemizethanks{\IEEEcompsocthanksitem L. Drumond was with the Information Systems and Machine Learning Lab, 
University of Hildesheim, Hildesheim, Germany.\protect\\
E-mail: ldrumond@ismll.de
\IEEEcompsocthanksitem Ernesto Diaz-Aviles was with IBM Research, Dublin, Ireland. \protect\\
E-mail: e.diaz-aviles@acm.org%
\IEEEcompsocthanksitem L. Schmidt-Thieme is with the Information Systems and Machine Learning Lab, 
University of Hildesheim, Hildesheim, Germany.\protect\\
E-mail: schmidt-thieme@ismll.de}
\thanks{Manuscript submitted April, 2016; }}

%

\maketitle

\IEEEtitleabstractindextext{%
\begin{abstract}
Learning from multiple-relational data which contains noise, ambiguities, or duplicate entities is essential to a wide range of applications such as statistical inference based on Web Linked Data, recommender systems, computational biology, and natural language processing.
These tasks usually require working with very large and complex datasets -- e.g., the Web graph -- however, current approaches to multi-relational learning are not practical for such scenarios due to their high computational complexity and poor scalability on large data.
In this paper, we propose a novel and scalable approach for multi-relational factorization based on consensus optimization. Our model, called ConsMRF, is based on the Alternating Direction Method of Multipliers (ADMM) framework, which enables us to optimize each target relation using a smaller set of parameters than the state-of-the-art competitors in this task.
Due to ADMM's nature, ConsMRF can be easily parallelized which makes it suitable for large multi-relational data. Experiments on large Web datasets --~derived from DBpedia, Wikipedia and YAGO~-- show the efficiency and performance improvement of ConsMRF over strong competitors. In addition, ConsMRF near-linear scalability indicates great potential to tackle Web-scale problem sizes. 
\end{abstract}

}

\IEEEdisplaynontitleabstractindextext

%
\IEEEpeerreviewmaketitle

\section{Introduction}
\label{sec:intro}
The complex graph structure of the Web -- with different relations or edge types -- has motivated a large body of research tackling the challenge of mining multi-relational data in the presence of noise, partial inconsistencies, ambiguities, or duplicate entities. State-of-the-art advances in this field are relevant to many applications such as link prediction~\cite{nickel:2012}, Resource Description Framework (RDF) mining~\cite{drumond:2012}, entity linking~\cite{Shen:2012:LLN:2187836.2187898}, recommender systems~\cite{artus:2012}, and natural language processing~\cite{jenatton:2012}. However, new paradigms are still needed for statistical and computational inference for very large multi-relational datasets, like the ones produced at massive scale in projects such as the Google's Knowledge Graph~\cite{Dong:2014:KVW:2623330.2623623}, YAGO~\cite{Suchanek:2007:YCS:1242572.1242667}, and in Semantic Web initiatives such as DBpedia~\cite{dbpedia:2007}.

Factorization models are considered state-of-the-art approaches for Statistical Relational Learning (SRL) in which they have exhibited a high predictive performance~\cite{drumond:2014,nickel:2015,singh:2008}. Factorization models for multi-relational data associate entities and relations with latent feature vectors and model predictions about unknown relationships through operations on these vectors (e.g., dot products). 
%
%

Optimizing the predictions for a number of relations can be seen as a prediction task with multiple target variables. For example, multi-target models can support information retrieval tasks in Linked Open Data bases like DBPedia by providing estimates of facts, that are neither explicitly stated in the knowledge base nor can be inferred from logical entailment, enabling probabilistic queries on such databases~\cite{nickel:2012,drumond:2012}. Another example in the context of social web recommender systems, is that such services are not only interested in recommending, for instance, news items to a user but also recommending other users as potential new friends.

State-of-the-art factorization models approach the multi-target prediction task by sharing the parameters used for all target relations. Instances of such approaches are RESCAL \cite{nickel:2012}, Multiple Order Factorization~\cite{jenatton:2012} and SME~\cite{glorot:2013}, which share entity specific parameters among all relations in the data. This way, the best solution for the optimization problem is a compromise of the performance on all relations. Although these models are  evaluated on multi-target settings, none of them explicitly investigates the problem of how to optimize each target relation \emph{individually} instead of learning the optimal performance compromise on all relations. The Decoupled Target Specific Features Multi-Target Factorization (DMF)~\cite{drumond:2014} addresses this drawback. 

DMF learns a set of models where each of them is optimized for a single \emph{target} relation and regularized by minimizing the loss on all the other (\emph{auxiliary}) relations in the data. The problem with this approach is twofold: (i)~the number of parameters grow too fast with the number of relations and (ii)~the runtime complexity is quadratic in the number of relations. The Coupled Auxiliary and Target Specific Features Multi-Target Factorization (CATSMF)~\cite{drumond:2014} alleviates the first problem by sharing a set of parameters among the models, but the second problem still persists. Besides, CATSMF needs to estimate a set of relation weights, a process that can be problematic -- e.g., setting the weights through model selection might be infeasible even for a moderate number of relations. Hence, CATSMF is still not scalable enough to handle large-scale multi-relational problems. 

In this paper, we propose \emph{ConsMRF} a novel approach for multi-relational factorization based on consensus optimization. ConsMRF defines target specific parameters and regularizes them against a \emph{global consensus variable} while its competitors, DMF and CATSMF, iterate over \emph{all} auxiliary relations for each target in the data. Thus \mbox{ConsMRF} has lower runtime costs while still featuring the predictive quality achieved by target specific models. In addition to that, and thanks to a learning algorithm based on the Alternating Direction Method of Multipliers (ADMM), the ConsMRF training can be parallelized -- in a shared memory or distributed environment -- allowing it to scale to large problems.

The main contributions of this paper are:
\begin{itemize}
 \item We propose ConsMRF, a novel approach for large scale multi-relational factorization. ConsMRF is based on consensus optimization, which optimizes each target relation specifically but more efficiently than state-of-the-art competitors;
 
 \item We propose an ADMM based learning algorithm for ConsMRF, which is amenable to parallelization, a key property to attain scalability;
 \\
 
 \item We conduct extensive experiments on real-world datasets derived from DBpedia, Wikipedia and YAGO, which demonstrate that ConsMRF achieves state-of-the-art predictive performance and that, at the same time, scales to large data.
\end{itemize}

\section{Background and Related Work} 
\label{sec:problem_formulation}
In this section we define the problem of multi-relational factorization, introduce state-of-the-art methods in this field, and position our model.

Throughout this paper we will use uppercase bold face letters like $\vect{A}$ to denote matrices, and lowercase boldface for vectors, e.g., $\vect{a}$. The i-th row  of a matrix $\vect{A}$ will be denoted as $\vect{a}_i$. Scalars will be denoted as non-boldface letters, e.g., $i, R$. Finally, we will denote sets as calligraphic letters like~$\E$.

\subsection{Multi-relational learning}
\label{subsec:probform}

Relational data comprise a set of $R \in \N$ relations among a set of entities $\E$. In this paper we assume all the relations to be binary, that is, we have relationships between a subject $s$ and an object $o$. The dataset for a given relation $r \in \{1, \ldots, R\}$ can be described as 
$$\D_r \subseteq \E \times \E \times \R \; .$$

A multi-relational model associates each relation $r \in \{1, \ldots, R\}$ with a prediction function $\hat y_r : \E \times \E \rightarrow \R$, which is characterized by a set of model parameters $\Theta$. Given some training data, the task is to find the set of parameters $\Theta$ for which the test error on previously unseen test data:

\scalebox{0.9}{
\parbox[c][][c]{\columnwidth}{%
\begin{align*}
  \text{error}((\D_r\test)_{r=1,\ldots,R}, (\hat y_r)_{r=1,\ldots,R}) :=  \frac{1}{R} \sum_{r=1}^R L_r(\D_r\test, \hat y_r(\cdot|\Theta))
\end{align*}
}
}
is minimal.

The loss function $L_r$ depends on the task and the nature of the data. For regression and classification problems the losses $L_r$ are usually defined as a sum of pointwise losses $\ell_r$:
\begin{align*}
   L_r(\D_r\test, \hat y_r) := \frac{1}{|\D_r\test|} \sum_{(s,o,y) \in \D_r\test} \ell_r(y, \hat y_r(s,o|\Theta)) ~ .
\end{align*}

Many multi-relational datasets however consist of positive-only instances, e.g., tuples of type $(s,o,1)$, like in Linked Open Data where only true triples are observed. In this case the standard setting is to optimize a pairwise ranking function:  
\begin{align*}
   L_r(\D_r\test, & \hat y_r) := \\ 
   & \frac{1}{|\D_r\test|} \sum_{(s,o,y) \in \D_r\test} \sum_{o' \in \Ob^r_s} \ell_r(\hat y_r(s,o|\Theta), \hat y_r(s,o'|\Theta)) ~ ,
\end{align*}
where $\Ob_s^r := \{o' \mid (s,o',y) \notin \D_r\}$ is the set of objects not linked to a given subject $s$ through relation $r$.

In total, the task at hand is an optimization problem that can be written as follows:
\begin{align*}
  \argmin_\Theta  \sum_{r=1}^R L_r(\D_r\train, \hat y_r(\cdot|\Theta)) + \Omega(\Theta) ~ ,
\end{align*}
where $\Omega$ is a regularization function. 

\subsection{Factorization models for multi-relational learning}

Factorization models define a matrix $\vect{A} \in \R^{|\E| \times k}$ where each of its rows, $\vect{a}_e \in \R^k$, is the  respective $k$-dimensional feature vector of entity $e \in  \E$. In addition, the models associate each relation $r$ with a matrix $\vect{W}_r \in \R^{k \times k}$. Thus the prediction corresponds to:
\begin{align}\label{eq:pred-func}
 \hat y_r(s,o) := \vect{a}_s^T \vect{W}_r\vect{a}_o ~ .
\end{align}

Most multi-relational factorization approaches differ by how they parametrize the relation feature matrices $\vect{W}_{r}$. Early models like the Collective Matrix Factorization (CMF) \cite{singh:2008} do not employ relation features, i.e., they can be viewed as defining $\W_r$ to be the $k \times k$ identity matrix. This may lead to poor prediction quality, specially because the prediction for different relations between the same pair of entities will be the same. To cope with this issue different approaches associate latent features with the relations.

The simplest approach that includes relation features is to define $\vect{W}_r$ as a diagonal matrix, a model that is equivalent to a PARAFAC tensor decomposition \cite{harshman:ucla70}. The Semantic Matching Energy (SME) model also uses this approach although with a slightly different prediction function~\cite{glorot:2013}. Using a full matrix for $\vect{W}_{r}$ is the approach used by RESCAL~\cite{nickel:2011}, the Multiple Order Factorization \mbox{(MOF)} \cite{jenatton:2012}, and the Localized Factor Model (LFM)~\cite{agarwal:2011}. Finally, approaches exist to deal with higher arity relations, e.g., the Coupled Matrix and Tensor Factorization (CMTF)~\cite{acar:2011} and MetaFac~\cite{lin:2009}. For the purposes of this work we focus on binary relations; however, the concepts described here can be easily applied to higher order relations. 

Note that none of the aforementioned state-of-the-art approaches make any distinction between target and auxiliary relations, and all of them use the same parameters for predicting all the targets, i.e. the learned parameters are a compromise for the performance over all targets, but not for each specific one. DMF, CATSMF, and more efficiently, ConsMRF  address this drawback.

\subsubsection*{DMF and CATSMF}
The Decoupled Target Specific Features Multi-Target Factorization (DMF) and the Coupled Auxiliary and Target Specific Features Multi-Target Factorization (CATSMF)~\cite{drumond:2014} combine the idea of shared parameters and learn individual entity embeddings for different target relations. DMF and CATSMF achieve state-of-the-art results for statistical relational learning tasks in comparison to RESCAL and MOF~\cite{drumond:2014}.

However, when learning a model with DMF the number of parameters grows too fast with the number of relations, an issue that CATSMF solves by sharing parameters among the models for different targets. In spite of that, CATSMF still has to estimate a set of relation weights (hyperparameters) which can be problematic -- e.g., setting them through model selection might be infeasible even for a moderate number of relations. Hence, CATSMF is still not scalable to efficiently handle large-scale multi-relational problems.

Our approach, ConsMRF, for multi-relational factorization is more efficient than its competitors, DMF and CATSMF, since it does not require to set relation weights as CATSMF and can be parallelized in a straightforward manner. 
%

Parallel and distributed algorithms for factorization models have been developed for single relation datasets, e.g. recommender systems. State-of-the-art approaches like NOMAD \cite{yun:2014} and DSGD \cite{gemulla:2011} (based on stochastic gradient descent), CCD++ \cite{yu:2012} (which parallelizes a coordinate descent algorithm), and DS-ADMM~\cite{yu:2014} (based on the ADMM) work well for problems like recommender systems where only one relation is available between two entity types --~i.e., user and items. Such strong assumptions on the data schema make these parallelization approaches not generalizable to the multi-relational case. 

ConsMRF, on the other hand, partitions the data relation-wise allowing for parallel processing on each of them. This property makes ConsMRF more attractive for applications that need to mine data with many relations. A recent work, Turbo-SMT~\cite{papalexakis:2014}, proposes to generate subsamples of the whole data, learn one model on each subsample, and combine them in a final step. Learning a model on each sample can be carried out in parallel. This framework however does not leverage target-specific features to improve prediction quality like ConsMRF does.

\subsection{Consensus Optimization and ADMM}

Consider the problem of minimizing a function \mbox{$f : \R^n \rightarrow \R$}. Assume that this function can be decomposed into $N$ components:
\begin{align*}
 f(\vect{x}) = \sum_{i=1}^N f_i(\vect{x})^{(t)}\enspace .
 \end{align*}

By defining local variables $\vect{x}_i$ for each component and a global variable $\vect{z}$, the minimization of $f$ can be reformulated as a consensus optimization problem \cite{bertsekas:1989}:
\begin{equation} 
\label{prob:consensus}
 \min_{\vect{z}, \{\vect{x}_i\}_{i = i, \ldots, N}}  \sum_{i=1}^N f_i(\vect{x}_i) \;\;\; \text{s.t.}  \;\;\; \vect{x}_i = \vect{z} \;\;\;\; i = 1, \ldots, N
\end{equation}
which can be solved via the Alternating Direction Method of Multipliers (ADMM)~\cite{boyd:2011} by minimizing the augmented Lagrangian of the problem:
\begin{equation}
 L(\{\vect{x}_i, \vect{v}_i\}_{i=1,\ldots,N},\vect{z}) = \sum_{i=1}^N f_i(\vect{x}_i) + \vect{v}_i^\top(\vect{x}_i-\vect{z}) + \frac{\rho}{2} ||\vect{x}_i-\vect{z}||_2^2 
\end{equation}
where $\vect{v}_i \in \R^n$ are the Lagrangian multipliers and $\rho > 0$ a penalty term.

ADMM is an iterative algorithm that works by finding the dual function 
$$g(\vect{V}) = \inf_{\vect{x}_i,\vect{z}}  L(\{\vect{x}_i\}, \vect{V},\vect{z})$$
and then maximizing $g$ in each iteration. The first step of finding 
$$ \inf_{\vect{x}_i,\vect{z}}  L(\{\vect{x}_i\},\vect{V},\vect{z})$$ 
is performed in an alternating fashion: in a first step the Lagrangian is minimized w.r.t. $\{\vect{x}_i\}_{i=1,\ldots,N}$, and then it is minimized w.r.t. $\vect{z}$. Thus for a given iteration $t$, the ADMM updates can be written as:
\begin{align*}
 \vect{x}_i^{(t+1)} &:= \argmin_\vect{x} f_i(\vect{x}) + {\vect{v}^{(t)}_i}^\top\vect{x} + \frac{\rho}{2} ||\vect{x}-\vect{z}^{(t)}||_2^2 \\
 \vect{z}^{(t+1)} &:= \argmin_\vect{z} \sum_{i=1}^N -{\vect{v}^{(t)}_i}^\top\vect{z} + \frac{\rho}{2} ||\vect{x}^{(t+1)}_i-\vect{z}||_2^2 \\
 \vect{v}_i^{(t+1)} &:= \vect{v}_i^{t} + \rho(\vect{x}^{(t+1)}_i-\vect{z}^{(t+1)}) \enspace .
\end{align*}

The ADMM algorithm has the appealing property that each $\vect{x}_i$ update can be performed in parallel. In addition, the z-update can be further simplified as follows. By deriving the objective function w.r.t. $\vect{z}$, making it equal to 0 and solving for $\vect{z}$, we arrive at the following analytical solution: 
\begin{align*}
 \vect{z}^{(t+1)} &:= \frac{1}{\rho N} \sum_{i=1}^N \vect{v}^{(t)}_i + \frac{1}{N}\sum_{i=1}^N \vect{x}^{(t+1)}_i   \enspace .
\end{align*}
If we choose the initial value $\vect{v}_i^{0} = \vect{0}$, then the values for $\vect{z}$ and $\vect{v}_i$ after the first iteration are:
\begin{align*}
 \vect{z}^{(1)} &:= \frac{1}{N} \sum_{i=1}^N \vect{x}_i^{(1)} \\ 
 \vect{v}_i^{(1)} &:= \rho \left(\vect{x}^{(1)}_i - \frac{1}{N} \sum_{j=1}^N \vect{x}_j^{(1)} \right) \enspace .
\end{align*}

On the second iteration, after updating $\vect{x}_i$, the value of $\vect{z}^{(2)}$  is:
\begin{align*}
 \vect{z}^{(2)} &:= \frac{1}{\rho N} \sum_{i=1}^N  \rho \left(\vect{x}^{(1)}_i - \frac{1}{N} \sum_{j=1}^N \vect{x}_j^{(1)} \right) + \frac{1}{N}\sum_{i=1}^N \vect{x}^{(2)}_i \\
 & = \frac{1}{N} \left( \sum_{i=1}^N \vect{x}^{(1)}_i -  \frac{1}{N}\sum_{i=1}^N \sum_{j=1}^N \vect{x}_j^{(1)} \right) + \frac{1}{N}\sum_{i=1}^N \vect{x}^{(2)}_i \\
 & = \frac{1}{N} \left( \sum_{i=1}^N \vect{x}^{(1)}_i -  \sum_{j=1}^N \vect{x}_j^{(1)} \right) + \frac{1}{N}\sum_{i=1}^N \vect{x}^{(2)}_i \\
 & = \frac{1}{N}\sum_{i=1}^N \vect{x}^{(2)}_i  \enspace .
\end{align*}

By repeating this procedure to subsequent iterations, it can be easily shown that, if we choose the initial value $\vect{v}_i^{0} = \vect{0}$, the z-update is reduced to: 
\begin{align*}
 \vect{z}^{(t+1)} := \frac{1}{N} \sum_{i=1}^N \vect{x}_i^{(t+1)} \enspace .
\end{align*}

Given its simplicity and power to solve distributed convex optimization problems, ADMM has recently found wide application in a number of areas in statistics and machine learning~--~e.g., matrix factorization~\cite{yu:2014}, tensor~\cite{6138863} and matrix completion~\cite{goldfarb2013}, control systems~\cite{6422363}, regression with hierarchical interactions~\cite{bien2013}, power systems~\cite{DBLP:journals/ftopt/KraningCLB14}, 
and computational advertising~\cite{Agarwal:2013:CAL:2541167.2514690}. 

However, to the best of our knowledge, ConsMRF is the first parallel algorithm for learning multi-relational factorization models under an ADMM framework. Our ADMM based learning algorithm enables ConsMRF not only to outperform the state-of-the-art competitors, DMF and CATSMF, but also to gracefully scale to large datasets.

\section{Optimizing for Multiple Relations}

In order to illustrate our approach, we introduce a running example used across the paper. Consider a social media website where users can follow other users (much like in Twitter), be friends with other users (forming a social graph) and consume products, e.g., read news items. In this example there are two entity types, namely users $U$ and news items $N$, and three relations: (i)~\emph{follows} $F$, (ii)~the \emph{social} relationship $S$ and (iii)~the product \emph{consumption} $C$, e.g., reading of news items. 
The task is, given existing past data, to recommend new friends, users to follow, and items to be consumed. 

Factorization models for this task are learned by optimizing the following function
\begin{align} \label{prob:shared}
  \argmin_{\A, \{\W_r\}_{r=1,\ldots,R} } \sum_{r=1}^R L_r(\D_r\train, \hat y_r(\cdot|\A, \W_r)) + \Omega(\A, \W_r) \enspace ,
\end{align}
which, in the example above, can be written as
\begin{align*} 
  \argmin_{\A, \W_S, \W_F, \W_C } & L_S(\D_S\train, \hat y_S(\cdot|\A, \W_S)) \\ 
  + &  L_F(\D_F\train, \hat y_F(\cdot|\A, \W_F)) \\
  + & L_C(\D_C\train, \hat y_C(\cdot|\A, \W_C)) \\ 
  + & \Omega(\A, \W_S, \W_F, \W_C) \enspace .
\end{align*}
However, this approach can at its best find model parameters whose performance are a compromise over all relations, as observed by Drumond et al.~\cite{drumond:2014}. 

A more suitable approach is to associate a different matrix $\A_r$ with each target relation so that the prediction functions can be optimized for each specific target. The naive approach would be to factorize each relation individually:
\begin{align*} 
  \argmin_{\A_S, \W_S} \enspace & L_S(\D_S\train, \hat y_S(\cdot|\A_S, \W_S)) + \Omega(\A_S, \W_S) \\
  \argmin_{\A_F, \W_F} \enspace & L_F(\D_F\train, \hat y_F(\cdot|\A_F, \W_F)) + \Omega(\A_F, \W_F) \\
  \argmin_{\A_C, \W_C} \enspace & L_C(\D_C\train, \hat y_C(\cdot|\A_C, \W_C)) + \Omega(\A_C, \W_C) \enspace .
\end{align*}

The problem with this approach is that the model learned for one relation does not exploit available information from the others. For instance, The social circle of a user as well as her taste (manifested through the items she consumed) are valuable predictors of whom she might be interested to follow. DMF alleviates that by optimizing the parameters for each relation $t$ over all the relations in the data thus solving the following problem~\cite{drumond:2014}:
\begin{align} 
\label{eq:dmf}
  \argmin_{\A, \{\W_{t,r}\}_{t,r=1,\ldots,R} } \sum_{t=1}^R\sum_{r=1}^R & \alpha_{t,r} L_t(\D_t\train, \hat y_r(\cdot|\A_t, \W_{t,r})) \\
  & + \Omega(\A_t, \W_{t,r}) \text{.} \nonumber
\end{align}

\begin{figure*}[!htb]
  \centering
  \subfigure[DMF]{
       \includegraphics[width=0.9\textwidth]{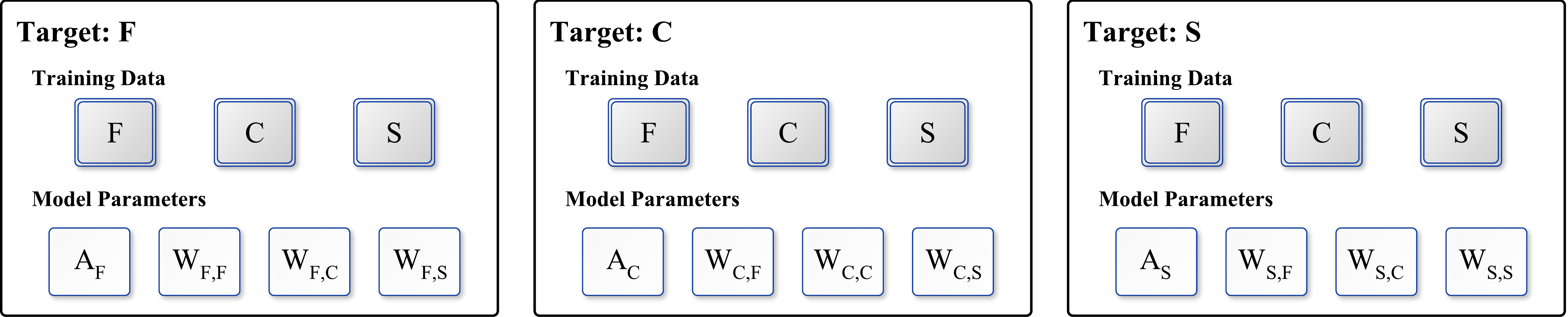}
       \label{fig:dmf-ex}
   }
       \subfigure[CATSMF]{
       \includegraphics[width=0.9\textwidth]{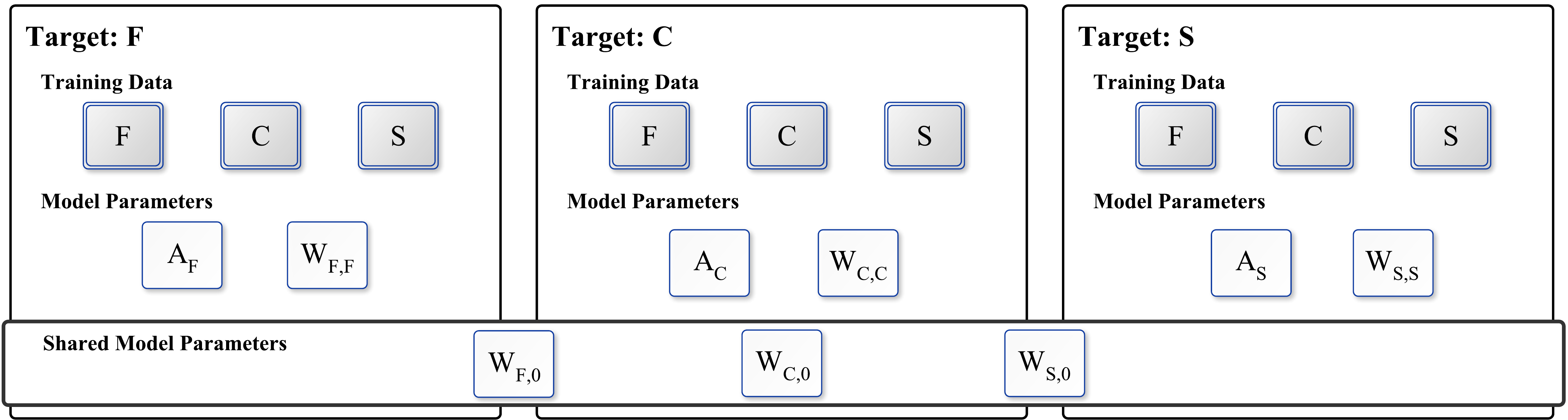}
       \label{fig:catsmf-ex}
   }
   \subfigure[ConsMRF]{
       \includegraphics[width=0.9\textwidth]{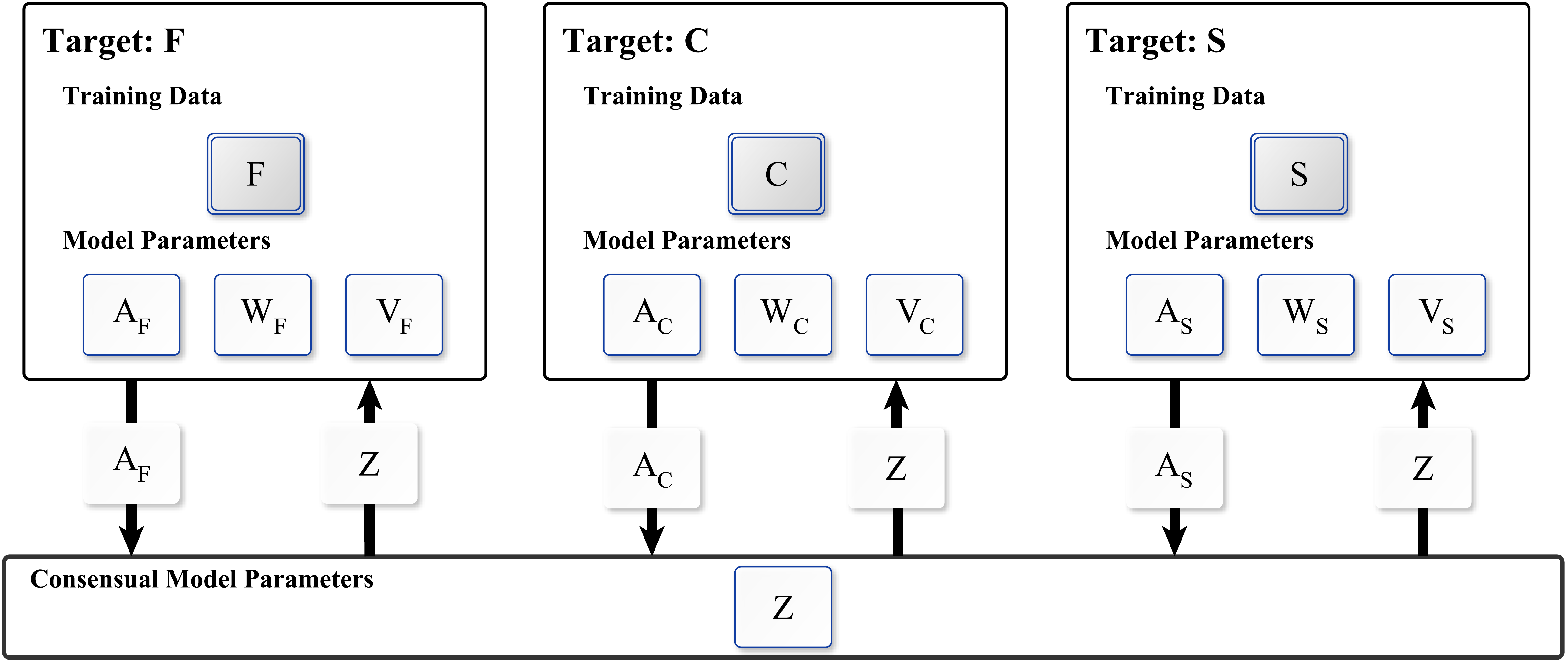}
       \label{fig:consmrf-ex}
   }
    \caption{DMF, CATSMF and ConsMRF on the social network example. The picture illustrates the data needed to optimize the parameters for each target relation. Gray boxes indicate available training data for each relation and white boxes, model parameters.}
    \label{fig:examples}
\end{figure*}

Figure \ref{fig:dmf-ex} illustrates how DMF would be applied for the social media website example. Note that the model for each target relation is learned on the whole data, and the model can be trivially parallelized as long as each worker has access to the \emph{whole} training data. One disadvantage of this approach is the additional amount of parameters needed. Figure~\ref{fig:catsmf-ex} shows how CATSMF \cite{drumond:2014} alleviates this issue by sharing redundant parameters. Observe that while CATSMF has less parameters than DMF, it is not easy to parallelize.

\section{Our Approach: The Consensus Multi-Relational Factorization}
\label{sec:approach}
In order to scale to large amounts of relational data without sacrificing predictive accuracy, an approach is needed that (i)~exploits target specific parameters for improved prediction accuracy and (ii)~can be efficiently parallelized and distributed. To address these two points we propose an approach based on the framework of \emph{consensus optimization}. 

Our new approach takes a different road than CATSMF and DMF: each model is learned only on the data about its target relation, while the parameters are regularized against a global consensus variable $\vect{Z}$. The advantages of this approach are three-fold; (i)~learning a model for a specific target relation is more efficient than DMF and CATSMF, since it only needs to iterate over the data of one relation; (ii)~the information about each relation flows through all the models by means of the variable $\vect{Z}$; (iii)~the model can be easily distributed by assigning each relation to a different machine (worker node), i.e., without requiring the duplication of the training data. 

We start from a model that employs solely target-specific parameters:
\begin{align} \label{eq:targ-spec}
   \argmin_{\{\A_r, \W_{r}\}_{r=1,\ldots,R} } \sum_{r=1}^R L_r(\D_r\train, \hat y_r(\cdot|\A_r, \W_{r})) + \Omega(\A_r, \W_{r}) \text{.} 
\end{align}
As discussed before, one strong disadvantage of this approach is that the parameters for predicting one relation are not learned exploiting the information about other relations. In order to alleviate this, we introduce one global entity feature matrix $\vect{Z}$:
\begin{align} 
\label{prob:cons}
 \min \;\; &\sum_{r=1}^R L_r\left(\D_r\train, \hat y_r(\cdot|\A_r, \W_r)\right) + \Omega(\A_r, \W_r) \\
 \text{s.t.}  \;\;\; &\A_r = \vect{Z} \;\;\;\; r = 1, \ldots, R  \enspace \nonumber
\end{align}
$\vect{Z}$ is called the ``consensus'' variable because it is used to make sure that the different $\A_r$ parameters converge to the same value. One consequence of this hard constraint is that the solution to problem~\ref{prob:cons} is equivalent to that of problem~\ref{prob:shared} thus, it cannot exploit fully the target specific parameters. We solve this by softening the constraints $\A_r = \vect{Z}$ and instead of solving problem~\ref{prob:cons}, we minimize its Lagrangian. This way, the \emph{Consensus Multi-Relational Factorization} (ConsMRF) problem can be formulated as:
\begin{align} \label{prob:consmrf}
 \sum_{r=1}^R & L_r\left( \D_r\train, \hat y_r \right) + \Omega(\A_r, \W_r) \\
 &+ {\vect{V}_r}^\top(\A_r - \vect{Z}) + \frac{\rho}{2} ||\A_r-\vect{Z}||_F^2 \enspace ~ \text{,} \nonumber
\end{align} 
%
where we denote $\hat y_r(\cdot|\A_r, \W_r)$ simply as $\hat y_r$ in order to avoid clutter. 

By minimizing the augmented Lagrangian, we can factorize each relation and still use information from the rest of the data by regularizing each local $\A_r$ against the global parameters $\vect{Z}$. 

$\A_r$, $\W_r$ and $\vect{Z}$ can be learned by solving problem~\ref{prob:consmrf} through the ADMM method. The ADMM algorithm for the Consensus Multi-Relational Factorization problem performs the following updates:
\begin{align} 
 \A_r^{(t+1)},\W_r^{(t+1)} &:= \argmin_{\A_r,\W_r}  L_r\left( \D_r\train, \hat y_r \right) + \Omega(\A_r, \W_r) \nonumber \\
 & +  {\vect{V}^{(t)}t_r}^\top\A_r + \frac{\rho}{2} ||\A_r-\vect{Z}^{(t)}||_F^2 \label{eq:updateAW}\\
 \vect{Z}^{(t+1)} &:=  \frac{1}{R} \sum_{i=1}^R \vect{A}_r^{(t+1)} \label{eq:updateZ}\\
 \vect{V}_r^{(t+1)} &:= \vect{V}_r^{t} + \rho(\A_r^{(t+1)}-\vect{Z}^{(t+1)}) \label{eq:updateV} \enspace .
\end{align}

While the updates for $\vect{Z}$ and $\vect{V}$ are clear and can be performed using a closed form solution, updating $\A_r$ and $\W_r$ involves performing an optimization task. Also these updates depend on the loss and regularization functions. In Section~\ref{subsec:update_A_and_W} we show what the updates for a pairwise loss function look like. 

\begin{algorithm}[!htb]
\caption{\textbf{~The ConsMRF Learning Algorithm}}
\scalebox{0.9}{
\parbox[c][][c]{\columnwidth}{%
\begin{algorithmic}[1]
\Procedure{LearnConsMRF}{}
\Statex \textbf{input: } number of relations $R$, training data $\{\D_r\}_{r = 1,\ldots,R}$, learning rate $\eta$,
\Statex regularization constant $\lambda$, and penalty~term~$\rho$
	\Statex
        \State $\vect{Z} \sim \mathcal{N}(0,\sigma^2\mathbf{I})$
        \Statex    
        \For {$r = 1,\ldots,R$}	  
          \State $\A_r \sim \mathcal{N}(0,\sigma^2\mathbf{I})$
	  \State $\W_r \sim \mathcal{N}(0,\sigma^2\mathbf{I})$
	  \State $\vect{V}_r \leftarrow \vect{0}$	  	
	\EndFor
	\Statex
	\Repeat
	  \PFor {$r = 1,\ldots,R$}  
	    \State $\A_r \leftarrow \vect{Z}$
	    \State $\A_r,\W_r \leftarrow \text{UpdateAW}(\D_r,\eta,\lambda, \A_r, \W_r, \vect{Z}, \vect{V}_r )$ 
	  \EndPFor
	  \Statex
	  \State $\vect{Z} \leftarrow  \frac{1}{R} \sum_{i=1}^R \vect{A}_r$
	  \Statex
	  \PFor {$r = 1,\ldots,R$}	  
	    \State $\vect{V}_r \leftarrow \vect{V}_r + \rho(\A_r-\vect{Z})$ 
	  \EndPFor	  
	\Until{convergence}	
\EndProcedure
\end{algorithmic}
}}
\label{alg:general}
\end{algorithm}

Algorithm~\ref{alg:general} summarizes the whole process and Figure~\ref{fig:consmrf-ex} illustrates it for our running example. After initializing the parameters, the entity and relation latent factors, $\A_r$ and $\W_r$, are updated using Stochastic Gradient Descent (SGD) as specified in Algorithm~\ref{alg:upaw}. It is important to state that the update for each relation $r$ is independent of the update for any other relation $s \neq r$. Thus, the \emph{for loop} in line 9 can be easily parallelized. On top of that, since each parallel worker only processes its own portion of the data, $\D_r$, and updates only its local variables, $\A_r$ and $\W_r$, this algorithm can be implemented both in a shared memory or in a distributed environment, provided that the complete data for one relation resides in one worker. Then, the algorithm  performs the updates on $\vect{Z}$ and $\vect{V}$ as in Equation~\ref{eq:updateZ} and Equation~\ref{eq:updateV}, respectively. 

DMF can also be parallelized, however, its ad-hoc parallelization is very inefficient when dealing with a high number of relations as observed in our empirical evaluation.

While a number of convergence criteria have been suggested for ADMM algorithms \cite{boyd:2011}, we have found out empirically that an early stopping strategy by checking the performance on the training set works well for our problem so that the algorithm can terminate if
\scalebox{0.88}{
\parbox[c][][c]{\columnwidth}{%
\[
 \left\lvert \sum_{r=1}^R L_r\left(\D_r\train, \hat y_r(\cdot|\A_r^{t}, \W_r^{t})\right) 
 - L_r\left(\D_r\train, \hat y_r(\cdot|\A_r^{t-1}, \W_r^{t-1})\right) \right\rvert  < \epsilon \text{.}
\]
}
}

To better understand the model, consider the social media example presented before. As it can be seen in Figure~\ref{fig:consmrf-ex}, the model can be learned by assigning one worker to each relation. In a first step, each worker optimizes locally and in parallel the local $\A_r$ and $\W_r$ variables. This means that the following problems are solved in parallel:
\begin{align*} 
  \argmin_{\A_S, \W_{S}} \enspace & L_S(\D_S\train, \hat y_S(\cdot|\A_S, \W_S)) + \Omega(\A_S, \W_S)  \\
  &+  \vect{V}^{(S)\top}(\A_S - \vect{Z}) + \frac{\rho}{2} ||\A_S-\vect{Z}||_F^2 \\
  \argmin_{\A_F, \W_{F}} \enspace & L_F(\D_F\train, \hat y_F(\cdot|\A_F, \W_F)) + \Omega(\A_F, \W_F) \\
  &+  \vect{V}^{(F)\top}(\A_F - \vect{Z}) + \frac{\rho}{2} ||\A_F-\vect{Z}||_F^2 \\
  \argmin_{\A_C, \W_{C}} \enspace & L_C(\D_C\train, \hat y_C(\cdot|\A_C, \W_C)) + \Omega(\A_C, \W_C) \\
  &+  \vect{V}^{(C)\top}(\A_C - \vect{Z}) + \frac{\rho}{2} ||\A_C-\vect{Z}||_F^2\enspace .
\end{align*}
Next, a centralized \emph{driver} node responsible for maintaining the consensus variable gathers the values of $\A_S$, $\A_C$ and $\A_F$ and updates $\vect{Z}$:
\begin{align*}
 \vect{Z} :=  \frac{1}{3} \left( \A_S+\A_C+\A_F \right) \text{.}
\end{align*}
Finally, the updated $\vect{Z}$ is broadcasted so that a new iteration can start. Note that this approach avoids inefficient data duplication in a distributed setting. 

Next we will see how the updates on each local worker are performed.

\subsection{Updating $\A_r$ and $\W_r$}
\label{subsec:update_A_and_W}
In each iteration, updating $\A_r$ and $\W_r$ requires solving the optimization problem given by Equation \ref{eq:updateAW}. To avoid overfitting we regularize the parameters using L2-regularization:
\begin{align*}
 \Omega(\A_r, \W_r) := \lambda (|| \A_r||_F^2 + ||\W_r||_F^2) \enspace ,
\end{align*}
where $||\cdot||_F$ denotes the Frobenius norm, so that our problem now is, for each relation $r$, to find the following latent features:
\begin{align*} 
\argmin_{\A_r,\W_r} \enspace &L_r\left( \D_r\train, \hat y_r \right) + \lambda (|| \A_r||_F^2 + ||\W_r||_F^2) \\ 
 &+  {\vect{V}^{(t)}_r}^\top\A_r + \frac{\rho}{2} ||\A_r-\vect{Z}^{(t)}||_F^2  \enspace . 
\end{align*}

Notice that in this framework one can optimize the relational model for a variety of loss functions, which ideally should approximate the evaluation criterion. Since most relational learning problems are evaluated using ranking measures, it is reasonable to optimize the models for a pairwise ranking function. To this end, we use the BPR optimization criterion (BPR-Opt) proposed by \cite{rendle-bpr:2009}. 

BPR-Opt is a smooth approximation of the Area Under the ROC Curve (AUC), thus enabling AUC optimization through standard gradient-based approaches. Also previous work has provided empirical evidence that it is an effective optimization criterion for the task approached here~\cite{drumond:2012}.

Let $\sigma(x) = \frac{1}{1+e^{-x}}$ denote the sigmoid function, BPR-Opt is an instance of a pairwise loss that can be defined for a general multi-relational learning task as follows:

\scalebox{0.88}{
\parbox[c][][c]{\columnwidth}{%
\begin{align*}
L_r^\text{BPR}\left( \D_r\train, \hat y_r \right) := - \sum_{(s,o,y) \in D_r\test} \sum_{o' \in \Ob^r_s} \ln\sigma(\hat y_r(s,o) - \hat y_r(s,o')) \enspace .
\end{align*}
}
}

Since it is not feasible to find a closed form solution to the aforementioned problem, we resort to an approximate solution by means of SGD, which has proven to scale gracefully to large datasets~\cite{bottou-2010}. The procedure is depicted in Algorithm~\ref{alg:upaw}.

\begin{algorithm}[!t]
\caption{\textbf{~}}
\scalebox{0.79}{
\parbox[c][][c]{\columnwidth}{%
\begin{algorithmic}[1]
\Procedure{UpdateAW}{}
\Statex \textbf{input: } relation $r$, training data $\D_r$, learning rate $\eta$, and regularization constant $\lambda$, penalty term $\rho$, latent features $\A_r, \W_r, \vect{Z}, \vect{V}_r$
\Statex \textbf{output: } updated latent features $\A_r, \W_r$
	\Statex
 	\Repeat
	  \State Draw $(s,o,y) \sim \D_r$
	  \State Draw $o' \sim \Ob^r_s$
	  \State $\vect{a_r}_s \leftarrow \vect{a_r}_s - \eta \left(\frac{\partial}{\partial\vect{a_r}_s}\ell_r(s,o,o') + \lambda \vect{a_r}_s + \vect{v_r}_s + \rho(\vect{a_r}_s-\vect{z}_s) \right)$
	  \Statex
	  \State $\vect{a_r}_o \leftarrow \vect{a_r}_o - \eta \left(\frac{\partial}{\partial\vect{a_r}_o}\ell_r(s,o,o') + \lambda \vect{a_r}_o + \vect{v_r}_o + \rho(\vect{a_r}_o-\vect{z}_o) \right)$
	  \Statex
	  \State \mbox{$\vect{a_r}_{o'} \leftarrow \vect{a_r}_{o'} - \eta \left(\frac{\partial}{\partial\vect{a_r}_{o'}}\ell_r(s,o,o') + \lambda \vect{a_r}_{o'} + \vect{v_r}_{o'} + \rho(\vect{a_r}_{o'}-\vect{z}_{o'}) \right)$} 
	  \Statex
	  \State $\W_r \leftarrow \W_r - \eta \left(\frac{\partial}{\partial\vect{W}_r}\ell_r(s,o,o') + \lambda \W_r \right) $
	\Until{convergence}	
	\State \Return $\A_r, \W_r$
\EndProcedure
\end{algorithmic}
}
}
\label{alg:upaw}
\end{algorithm} 

The algorithm starts by randomly sampling one observed data point, $(s,o,y) \in \D_r$, uniformly at random and one additional object $o' \in \Ob^r_s$. Recall the definition of $\Ob^r_s$ from Section \ref{sec:problem_formulation}, which is the set of objects not associated with subject $s$ through relation $r$, i.e.,~$\Ob_s^r := \{o' | (s,o',y) \notin \D_r\}$. 

Sampling objects from this set is performed as follows: an object $o'$ is sampled uniformly at random from $\E$ and it gets accepted if $ (s,o',y) \notin \D_r$ or another sample is drawn otherwise. Once $(s,o,y)$ and $o'$ are sampled, the algorithm continues by updating the respective parameters in the opposite direction of the gradient. 
%

For the BPR-Opt optimization the stochastic gradients (evaluated on only one data point) correspond to:
\scalebox{0.92}{
\parbox[c][][c]{\columnwidth}{%
\begin{align*}
  \frac{\partial \ell_r(s,o,o')}{\partial\theta} = 
  \begin{cases} 
  \frac{-1}{1 + e^{\hat y_r(s,o) - \hat y_r(s,o')}} \vect{W}_r(\vect{a_r}_o - \vect{a_r}_{o'}) &\mbox{if } \theta = \vect{a_r}_s \enspace , \\ 
  \frac{-1}{1 + e^{\hat y_r(s,o) - \hat y_r(s,o')}} \vect{a_r}_s\vect{W}_r &\mbox{if } \theta = \vect{a_r}_o \enspace ,\\ 
  \frac{1}{1 + e^{\hat y_r(s,o) - \hat y_r(s,o')}} \vect{a_r}_s\vect{W}_r &\mbox{if } \theta = \vect{a_r}_{o'} \enspace ,\\ 
  \frac{-1}{1 + e^{\hat y_r(s,o) - \hat y_r(s,o')}} \vect{a_r}_s(\vect{a_r}_o - \vect{a_r}_{o'}) &\mbox{if } \theta = \vect{W}_r \enspace . \\ 
  \end{cases} 
\end{align*}
}}

\subsection{Relation of ConsMRF to CATSMF, DMF, and \break Other Models}

Traditional factorization models like RESCAL~\cite{nickel:2011} or CMF~\cite{singh:2008} define one entity feature matrix $\A$ and one relation feature matrix per relation $\W_r$. Those parameters are learned by optimizing a loss function like in Equation~\ref{prob:shared}, so that the learned latent features $\A$ are the ones which provide the best performance compromise over all relations in the data.

One alternative to this approach is to learn one set of specific features per target relation thus optimizing the following loss function:
 
\[
 \sum_{t=1}^R L_t(\D_t\train, \hat y_t(\cdot|\A_t, \W_t)) \enspace \text{.}
\]

As already stated, the drawback here is that the model for one relation $r$ cannot learn anything from the information on other relations $t \neq r$, which is the whole point of multi-matrix factorization. CATSMF and DMF~\cite{drumond:2014} solve this problem by using information about other relations by introducing a regularization term. Such a regularization term for a relation $t$ can be written as:

\[
  \sum_{r=1, r\neq t}^R \alpha_{t,r} L_r(\D_r\train, \hat y_{t,r}(\cdot|\A_t, \W_{t,r})) \enspace ,
\]
with $\alpha_{t,r}$ being regularization weights and $\W_{t,r}$ for $t \neq r$ a set of auxiliary parameters which are learned for regularization purposes but never used for making predictions on the test data. It has been shown that this strategy leads to better predictive performance~\cite{drumond:2014}.

ConsMRF takes a similar approach but implements it in a more efficient and principled manner. Building on the theory of consensus optimization and the ADMM method, ConsMRF defines a global entity feature matrix $\vect{Z}$ and regularizes the parameters for a given relation $t$ using the following term:

\[
 \frac{\rho}{2} ||\A_t - \vect{Z}||_F^2 \enspace \text{,}
\]
where $\vect{Z}$ encodes the information of the other relations as it can be seen in Equation~\ref{eq:updateZ}. ConsMRF approach has two major advantages over the one of DMF and CATSMF: (i)~it does not involve two nested summations over the auxiliary relations; therefore, it scales better with the number of relations in the data, and (ii)~it avoids potentially cumbersome setting of the $\alpha_{t,r}$ hyperparameters (whose number grows quadratically on the number of relations). 

Finally, when resorting to parallelization, ConsMRF is also more efficient. The main reason is because, for DMF, each worker needs to access all the data. This might not be a problem in a shared memory setting but, in a distributed environment, the whole data needs to be replicated on each node. On the other hand, ConsMRF only requires that each worker has access to the data about the relation it is assigned to, thus no data duplication is necessary.

%

\section{Experimental Evaluation}
\label{sec:evaluation}
%
In this section, we assess the behavior of ConsMRF on practical Web applications in terms of predictive performance and scalability. We compare \mbox{ConsMRF} against the state-of-the-art competitors: DMF~\cite{drumond:2014}, CATSMF~\cite{drumond:2014},  RESCAL~\cite{nickel:2011}, and a standard Canonical Decomposition (CD)~\cite{harshman:ucla70}. We first describe the datasets, then the protocol and experimental setting, and conclude the section with the results and discussion of our empirical study.
 
\subsection{Datasets}
In our experiments we used three Web datasets collected from DBpedia, Wikipedia and YAGO, whose statistics are summarized in Table~\ref{tag:datasets_stats}. The datasets are described as follows.
 
\noindent\textbf{DBpedia}~\cite{dbpedia:2007} is one of the central interlinking-hubs of the emerging Web of Data,\footnote{\url{http://lod-cloud.net/}} which makes it really attractive to evaluate multi-relational learning approaches. Our dataset comprises 625,680 triples from a sample of the \textit{DBpedia Properties} in English\footnote{\url{http://downloads.dbpedia.org/3.6/}}. It contains 269,862 entities and 5 relations regarding the music domain namely \texttt{artist}, \texttt{genre}, \texttt{composer}, \texttt{associated\_band}, and \texttt{associated\_musical\_artist}.
 
\noindent\textbf{Wikipedia-SVO}~\cite{jenatton:2012} has one of the highest number of relations among benchmark datasets for multi-relational tasks. It contains subject-verb-object triples extracted from over two million Wikipedia articles, where the verbs play the role of the relationship. It consists of 1,300,000 triples, about 4,538 relations, and 30,492 entities.
 
\noindent\textbf{YAGO}~\cite{Suchanek:2007:YCS:1242572.1242667} is a huge semantic knowledge base derived from Wikipedia, WordNet\footnote{\url{https://wordnet.princeton.edu/}} and GeoNames\footnote{\url{http://www.geonames.org/}}. This dataset is made of the core facts of YAGO~2 ontology, i.e., the \texttt{yagoFacts} triples\footnote{\url{http://www.mpi-inf.mpg.de/yago/}}, which amount to 4,431,523 observations, 2,137,469 entities, and 37 relations.

\begin{table}[!tb]
\setlength{\tabcolsep}{10pt}
\centering
\caption{Datasets statistics.}
\begin{tabular}{l r r r}
\toprule
\textbf{Dataset}       & Entities & Relations & Triples \\
\midrule
\textbf{DBpedia}       & 269,862  & 5        & 625,680 \\
\textbf{Wikipedia-SVO} & 30,492 & 4538 & 1,300,000 \\
\textbf{YAGO}          & 2,137,469 & 37      & 4,431,523 \\
\bottomrule
\end{tabular}
\label{tag:datasets_stats}
\end{table}

\subsection{Evaluation Protocol and Experimental Settings}
The dataset is split into training, validation, and test set. First, we randomly select 10\% of the positive tuples and assign them to the test set. Then, we randomly sample 10\% of the remaining positive tuples to form the validation set. The rest of the tuples are used for training. 

To account for variability, we perform a 10-fold cross-validation. The results reported in Figure \ref{fig:resultsbig} are the average over the rounds, while the error bars represent 99\% confidence intervals. 

We expect a good model to score true facts higher than the false ones (i.e., unobserved), thus we are dealing with a ranking task, which leads us to the following evaluation protocol based on~\cite{koren:2010}. For each relation $r$ and entity $s$ on the test set:
\begin{enumerate}
  \item First, we sample a set of unobserved triples in the knowledge base, i.e., 
  $$\mathcal{R}_{r,s}^- \subseteq \{(s,o',0) | (s,o',1) \notin \D_r\train \cup \D_r\val \cup \D_r\test  \} \; .$$  This sampling is performed by drawing an object $o' \approx \E$ and if $ (s,o',1) \notin \D_r\train \cup \D_r\val \cup \D_r\test$, the triple $(s,o',0)$ is added to $\mathcal{R}_{r,s}^-$, otherwise another sample is drawn.
  
  \item Then, we compute the score for this sample of unobserved triples, $\mathcal{R}_{r,s}^-$, as well as  for each of the observed ones in the test set 
  $$\mathcal{R}_{r,s}^+ = \{(s,o,1) | (s,o,1) \in \D_r\test\} \; .$$

  \item Finally, we rank the triples $\mathcal{R}_{r,s}^+ \cup \mathcal{R}_{r,s}^-$ based on the models assessed in this empirical evaluation and their performance is measured by looking at the following metrics: \textbf{AUC} (area under the ROC curve), \textbf{precision} at $5$ and \textbf{recall} at $5$.
 \end{enumerate}
 
 The reported results are averaged over all relations $r$ and subjects $s$. 
 
 All experiments were executed on a GNU/Linux machine running CentOS version 6.5 equipped with an Intel Xeon-Phi E5-2670 2.50GHz (40 cores) processor and 128GB RAM. ConsMRF was implemented in C++ using the Eigen library~\cite{eigenweb} and OpenMP\footnote{\url{http://openmp.org/}} for parallelization. 
 
 For each split we also sampled one hold-out set used to tune hyperparameters. For all the models we searched the number of latent features in the range $k \in \{10, 25, 50\}$, except for Wikipedia-SVO where we used $k=10$ and the regularization weights $\lambda \in \{0.0005, 0.005, 0.05\}$. The learning rate, $\eta$, for ConsMRF, CATSMF and DMF was initialized with $\eta = 0.5$  and adjusted using the ADAGRAD policy~\cite{duchi:2011}. 

 For CATSMF and DMF the relation weights $\alpha_{t,r}$ were searched in the range $\alpha_{t,r} \in \{0, 0.25, 0.75, 1\}$. Due to the high number of relations in the \mbox{Wikipedia-SVO} and Yago dataset, we set each $\alpha_{t,r} = a$ and searched $a$ in the aforementioned interval as in \cite{drumond:2014}. Finally, for ConsMRF, the penalty parameter $\rho$ was searched in the range $\rho \in \{0.00005, 0.0005, 0.005\}$.

%
%
\subsection{Results and discussion}
\begin{figure*}[!t]
  \centering
  \subfigure[AUC]{
       \includegraphics[width=0.29\textwidth]{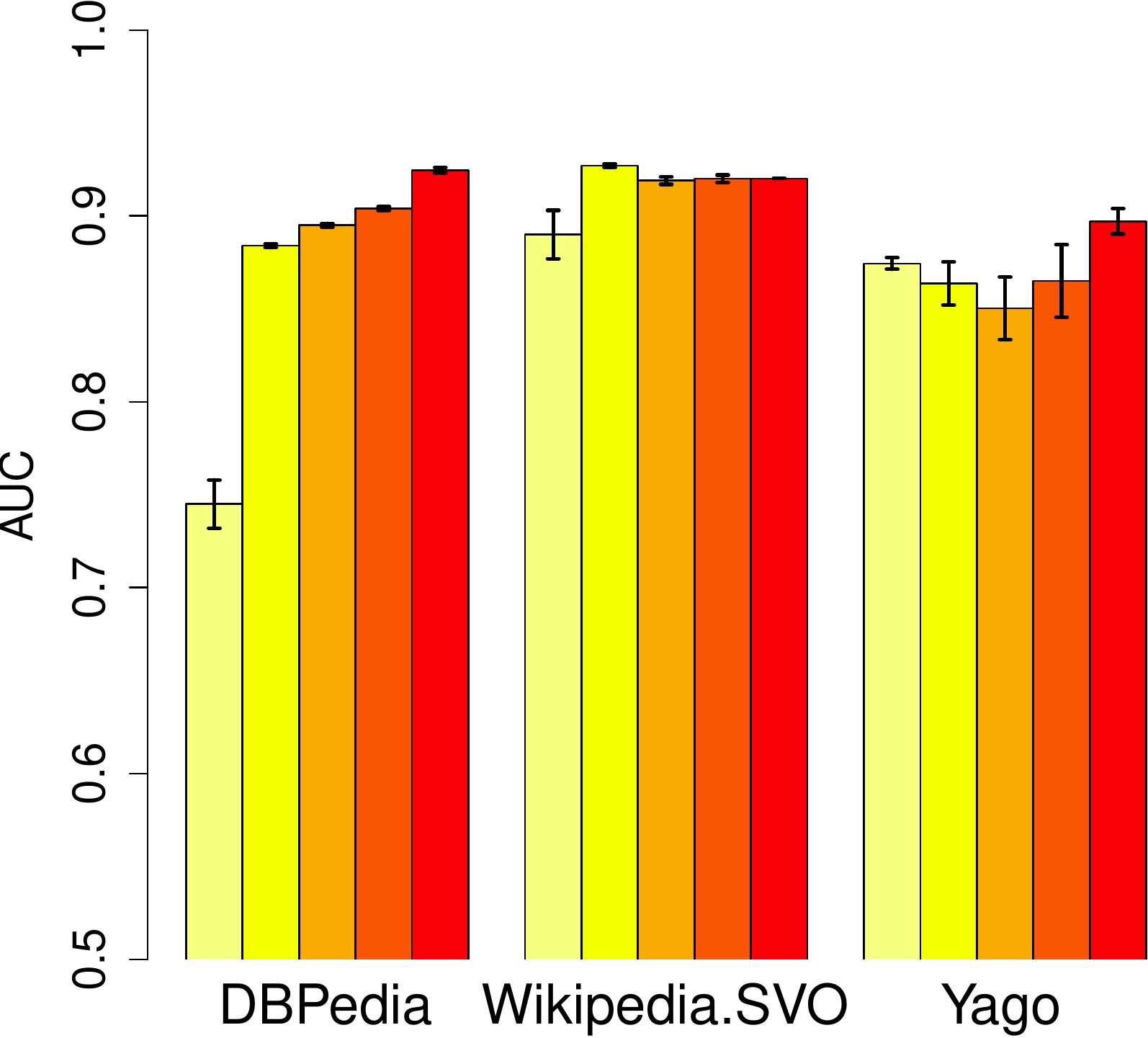}
       \label{fig:auc}
   }~
       \subfigure[Recall@5]{
       \includegraphics[width=0.29\textwidth]{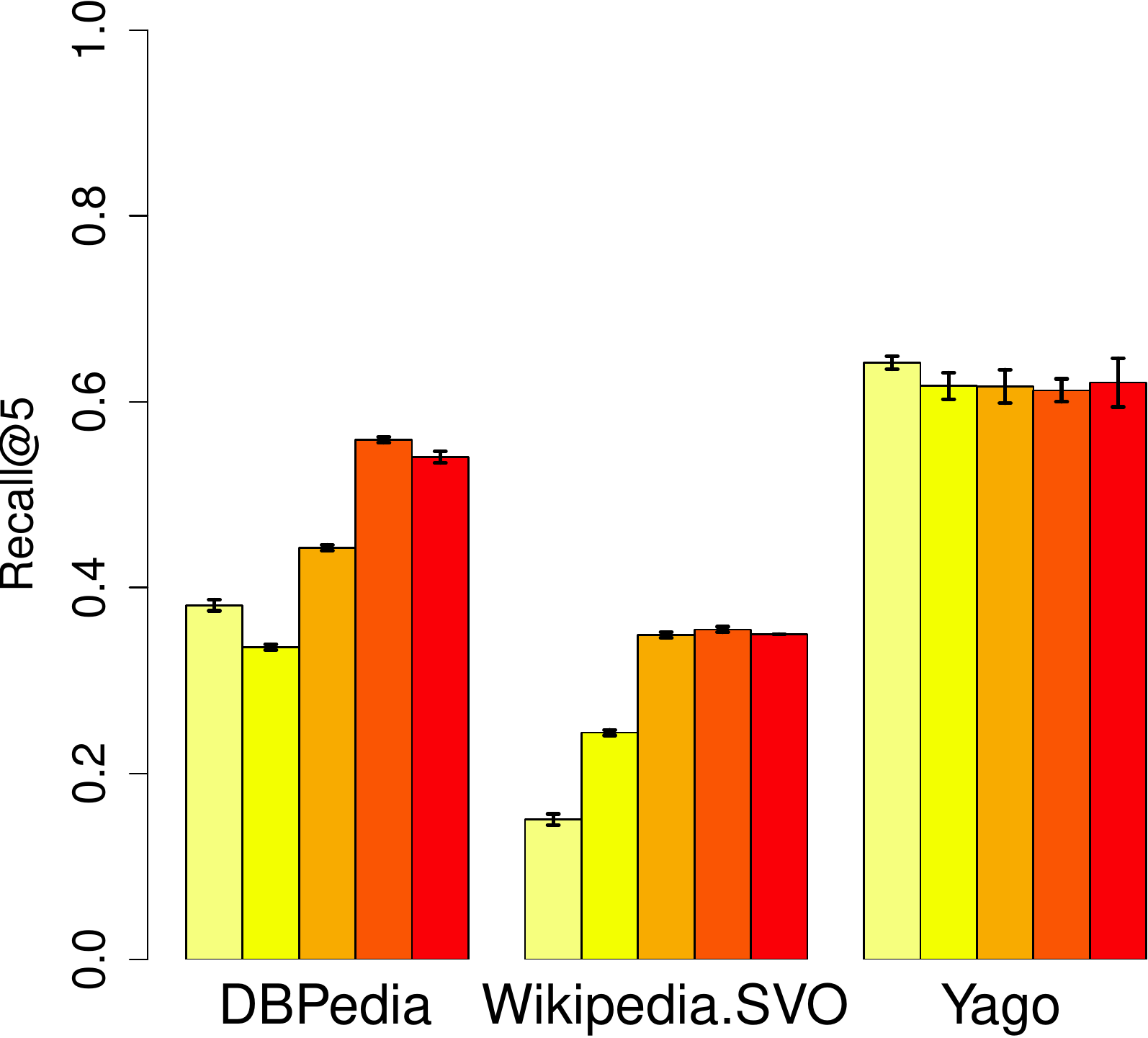}
       \label{fig:rec}

   }~
   \subfigure[Precision@5]{
       \includegraphics[width=0.29\textwidth]{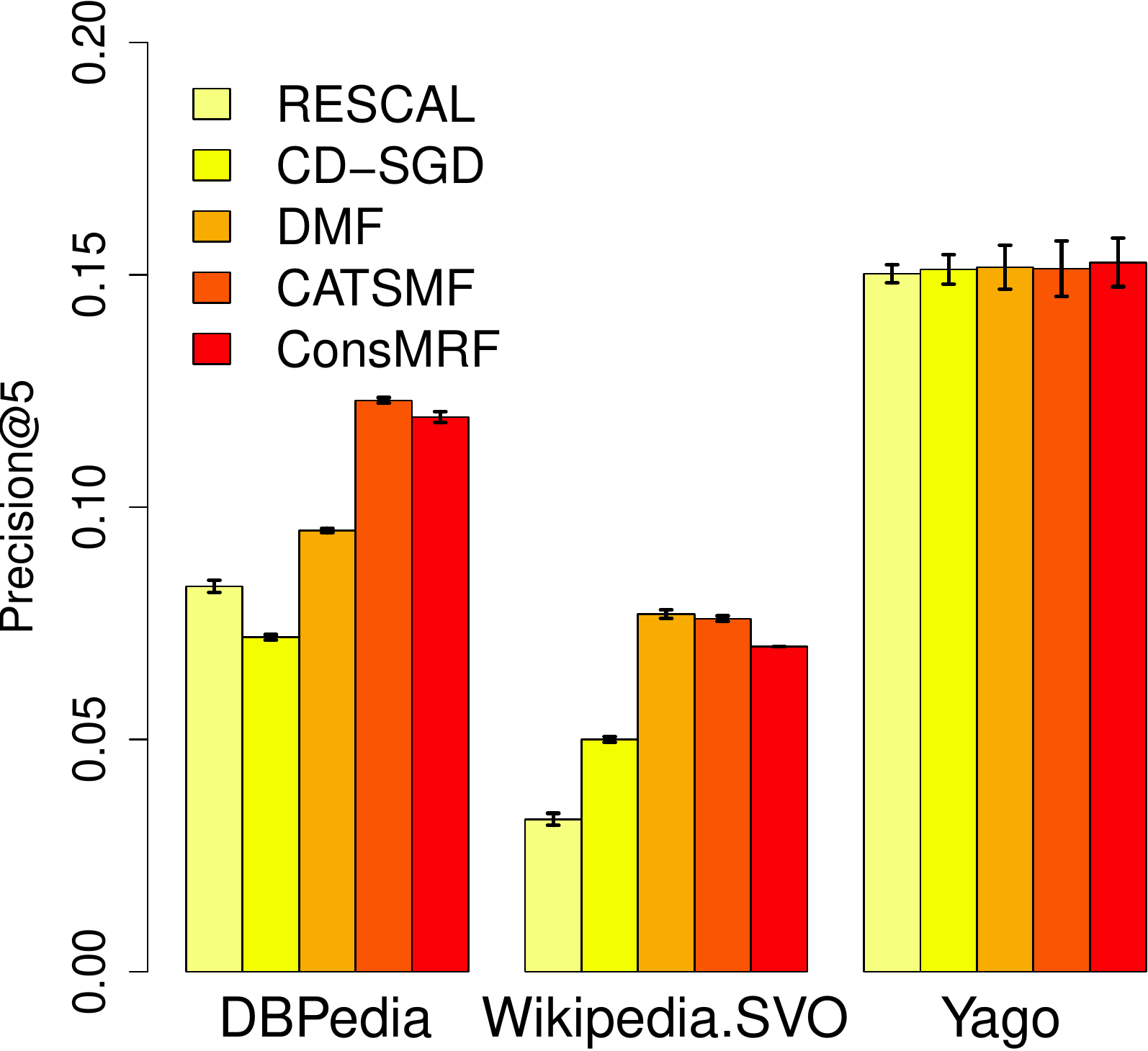}
       \label{fig:prec}
   }
    \caption{Experimental results on the DBpedia, Wikipedia-SVO, and YAGO datasets.}
    \label{fig:resultsbig}
\end{figure*}

In this evaluation we used diagonal $\W_r$ matrices in our ConsMRF model. We compare it against the DMF and CATSMF approaches also using diagonal matrices for relation features as well as a complete sharing approach, which is equivalent to a  standard Canonical Decomposition (CD) optimized with SGD. In all these models the relation loss used was the BPR-Opt, which approximates the AUC measure. To keep the results in perspective, we also added RESCAL to the evaluation, a well known state-of-the-art multi-relational factorization model. The results of our evaluation can be seen in Figure~\ref{fig:resultsbig}. 

\begin{table}[!tb]
    \centering
    \caption{Overall training runtime in seconds using one core.  Times are for a single run with fixed hyperparameters. ConsMRF speedup is computed w.r.t. DMF and CATSMF where `x' should be read as \emph{``times faster"}.} 
    \setlength{\tabcolsep}{5pt}
    \label{tab:runtime}
    \scalebox{0.85}{
    \begin{tabular}{l r r r c}
	\toprule
	\textbf{Dataset} & \textbf{DMF} & \textbf{CATSMF} & \textbf{ConsMRF} & \textbf{ConsMRF speedup over}\\ 
	{} & {} & {} & {} & \textbf{DMF~$/$~CATSMF}\\ 
	\midrule
	DBpedia              & 690s & 754s &  678s        &  0.02x $/$ 0.1x \\ 
	\mbox{Wikipedia-SVO} & 297,186s & 246,402s & 11,885s &  24x  $/$ 19.7x \\ 
	YAGO                 & 65,456s & 60,521s &  7,639s    &  8.5x  $/$ 7.9x  \\ 
	\bottomrule
    \end{tabular}    
    }
\end{table}

ConsMRF achieves the best AUC scores on all the datasets, with a tie on the Wikipedia-SVO. This is important because AUC is the measure all models are optimized for -- with the exception of RESCAL that is optimized for the L2-Loss. In other words, ConsMRF is able to achieve better scores on the measure it is optimized for, which is a promising result given that the framework is general enough to allow different loss functions, $L_r$, in Equation~\ref{prob:consmrf}.

ConsMRF performs slightly worse than CATSMF on precision (in DBpedia) and recall (in DBpedia and Wikipedia-SVO) but with the added advantage of being a parallel algorithm, thus being able to scale to larger datasets. Finally, unlike DMF and CATSMF, ConsMRF does not require careful tuning of the relation weights $\alpha_{t,r}$. 

%
\noindent\textbf{Scalability.~} To demonstrate ConsMRF ability to scale to large datasets we report runtime performance here. Table~\ref{tab:runtime} shows the total training time using \emph{only one core} for DMF, CATSMF and ConsMRF. 

We can observe that while the total runtime for all the methods is comparable on the DBpedia dataset (which consists of only five relations), ConsMRF has a much lower runtime for YAGO and Wikipedia-SVO. Note that the speedups for \mbox{Wikipedia-SVO} (4538 relations) are much higher than the speedups for YAGO (37 relations), confirming that ConsMRF scales gracefully with the number of relations unlike CATSMF and DMF.


Also observe that, while Table~\ref{tab:runtime} does not consider hyperparameter optimization, by setting $\alpha_{t,r} = a$, CATSMF and DMF have the same number of hyperparameters as ConsMRF.
%

\begin{figure*}[!tb]
  \centering
  \subfigure[Wikipedia-SVO]{
                  \includegraphics[width=0.29\textwidth]{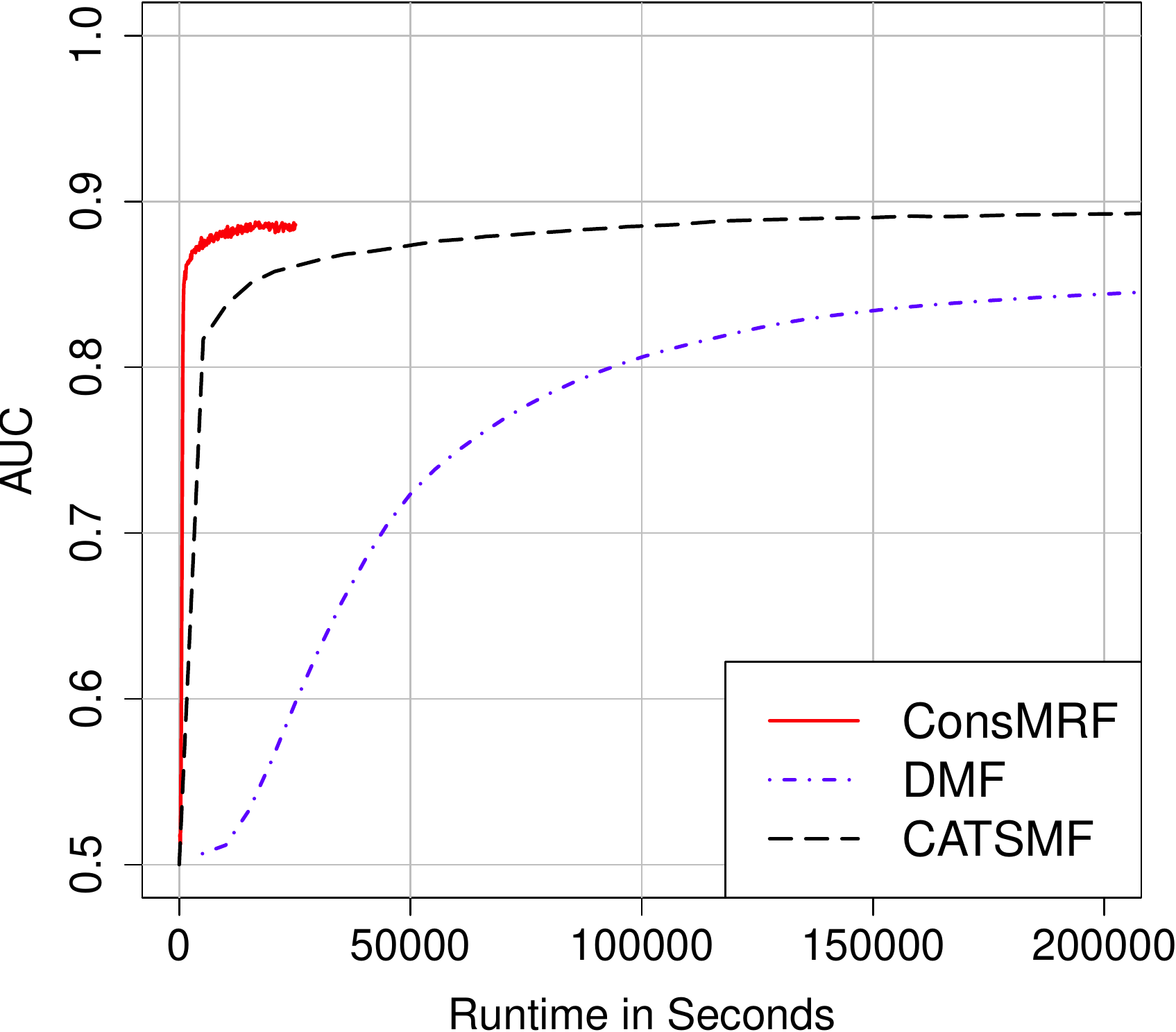}
                  \label{fig:convergence-svo}
  } ~
   \subfigure[DBpedia]{
                    \includegraphics[width=0.29\textwidth]{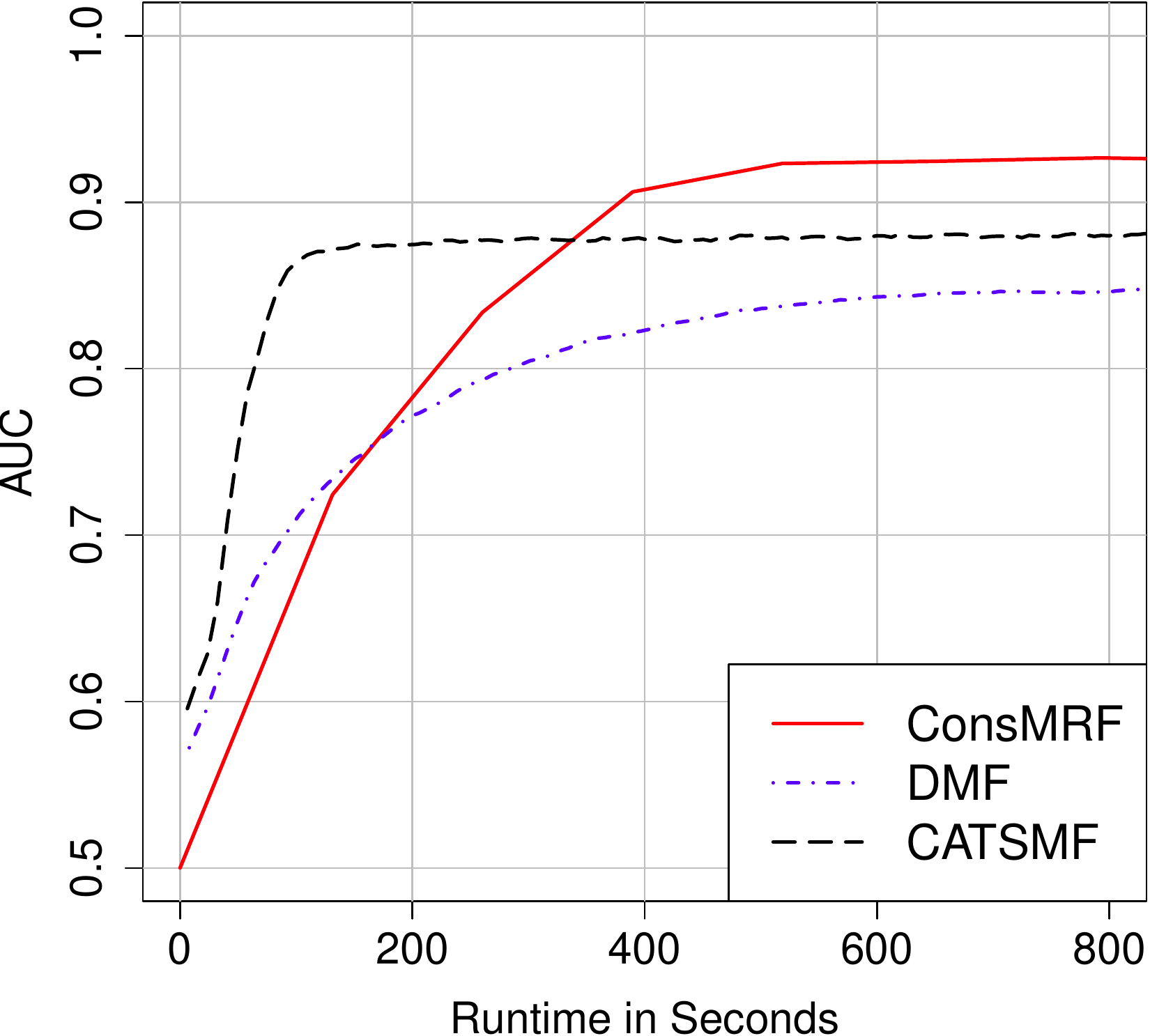}
                    \label{fig:convergence-db}
   } ~
   \subfigure[YAGO]{
                    \includegraphics[width=0.29\textwidth]{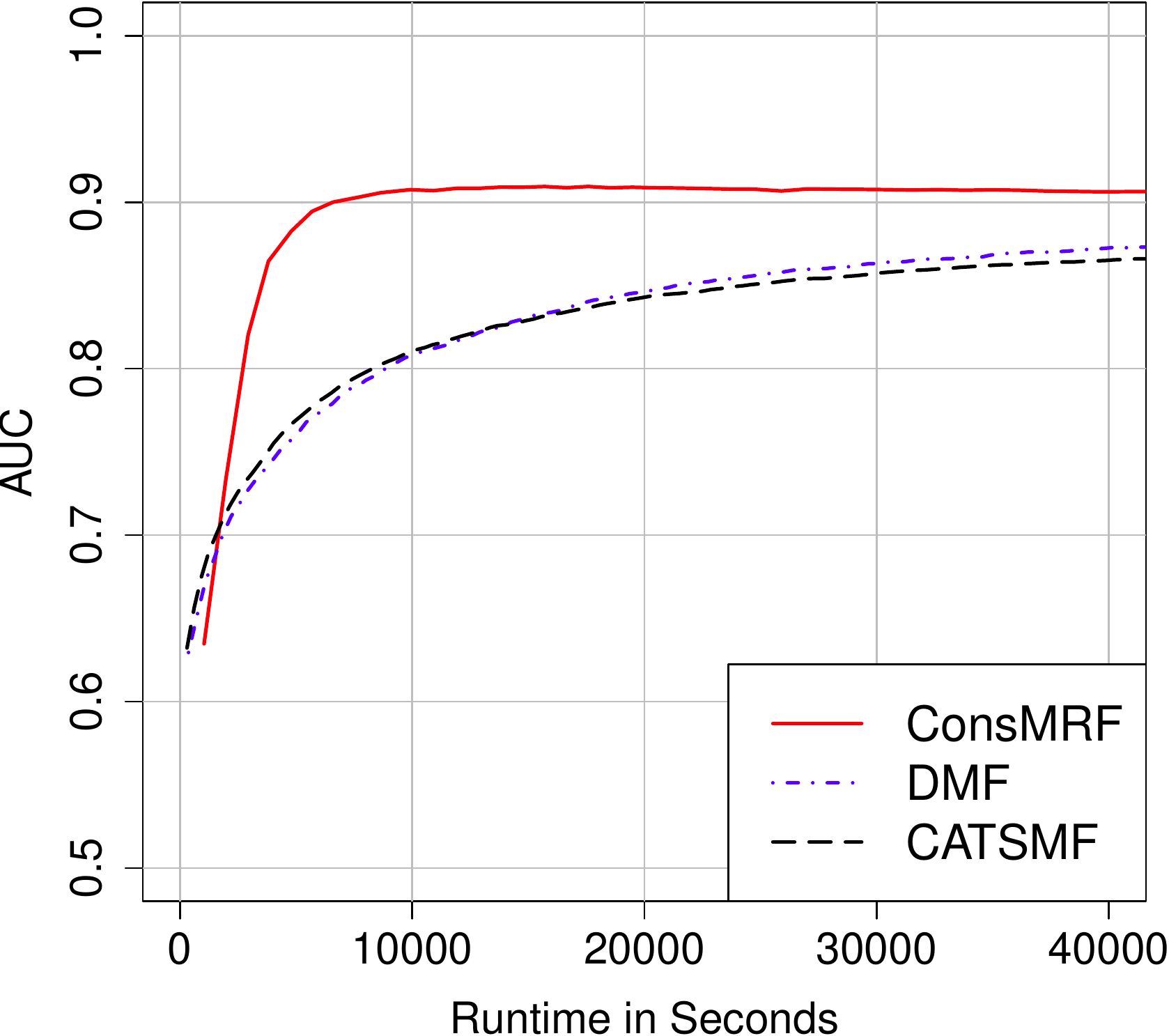}
                    \label{fig:convergence-yago}
   }
    \centering
    \caption{Learning curves showing the convergence speed of ConsMRF, DMF, and CATSMF using only one core.}
    \label{fig:convergence}
\end{figure*}

\begin{figure}[!tb]  
  \centering    
    \includegraphics[width=0.7\columnwidth]{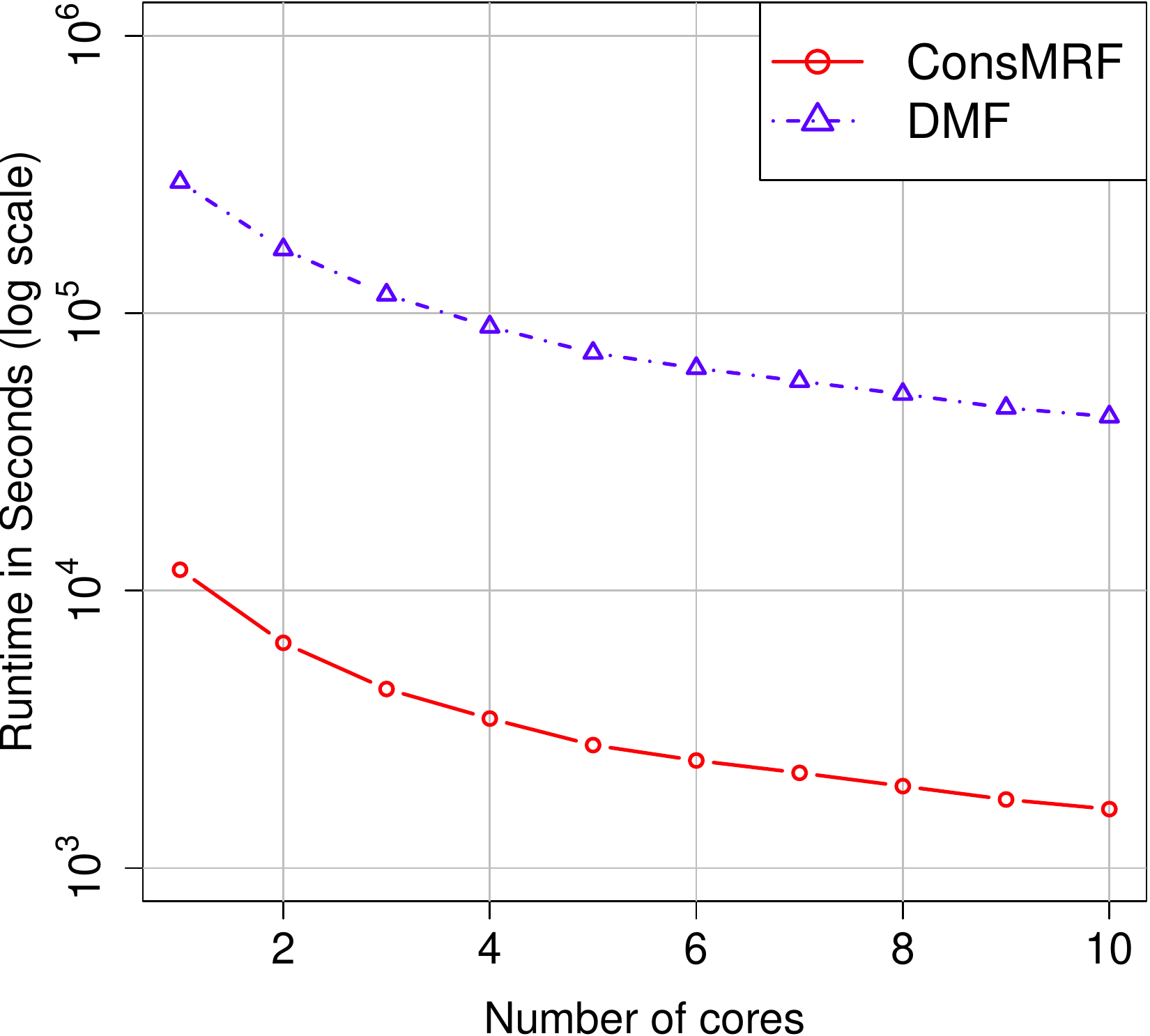}     
    \caption{Training wall-clock time in seconds against the number of cores on the \mbox{Wikipedia-SVO} dataset, which is the one with the largest number of relations.}
    \label{fig:cores}  
\end{figure}

A closer comparison of the runtime performance of ConsMRF, CATSMF and DMF can be seen on Figure~\ref{fig:convergence}. The figure shows the learning curves for both methods on the same machine using only one core. Note how ConsMRF converges much faster than CATSMF and DMF on the datasets with higher number of relations, namely \mbox{Wikipedia-SVO} (Figure~\ref{fig:convergence-svo}) and YAGO (Figure~\ref{fig:convergence-yago}). For the DBpedia dataset, which only has 5 relations, the convergence speed of the three methods is comparable (Figure~\ref{fig:convergence-db}). Finally, Figure~\ref{fig:cores} shows how ConsMRF and DMF scale with the number of cores. We plot the total training wall-clock time in seconds against the number of cores used. It is worth noting that the speedups achieved by ConsMRF are limited by two factors, namely: (i)~unbalanced workload and (ii)~synchronization costs. Given two distinct relations $r$ and $t$, if $|\D_r\train| > |\D_t\train|$ the node optimizing $\A_r, \W_r$, will have more work to do than the one learning $\A_t,\W_t$. Since the algorithm is synchronous some cores might need to wait for the others to finish their work. One possible way to avoid this problem would be to look into asynchronous ADMM approaches~\cite{zhang:2014}.

\begin{figure*}[!tb]
  \centering
  \subfigure[DBpedia]{
       \includegraphics[width=0.29\textwidth]{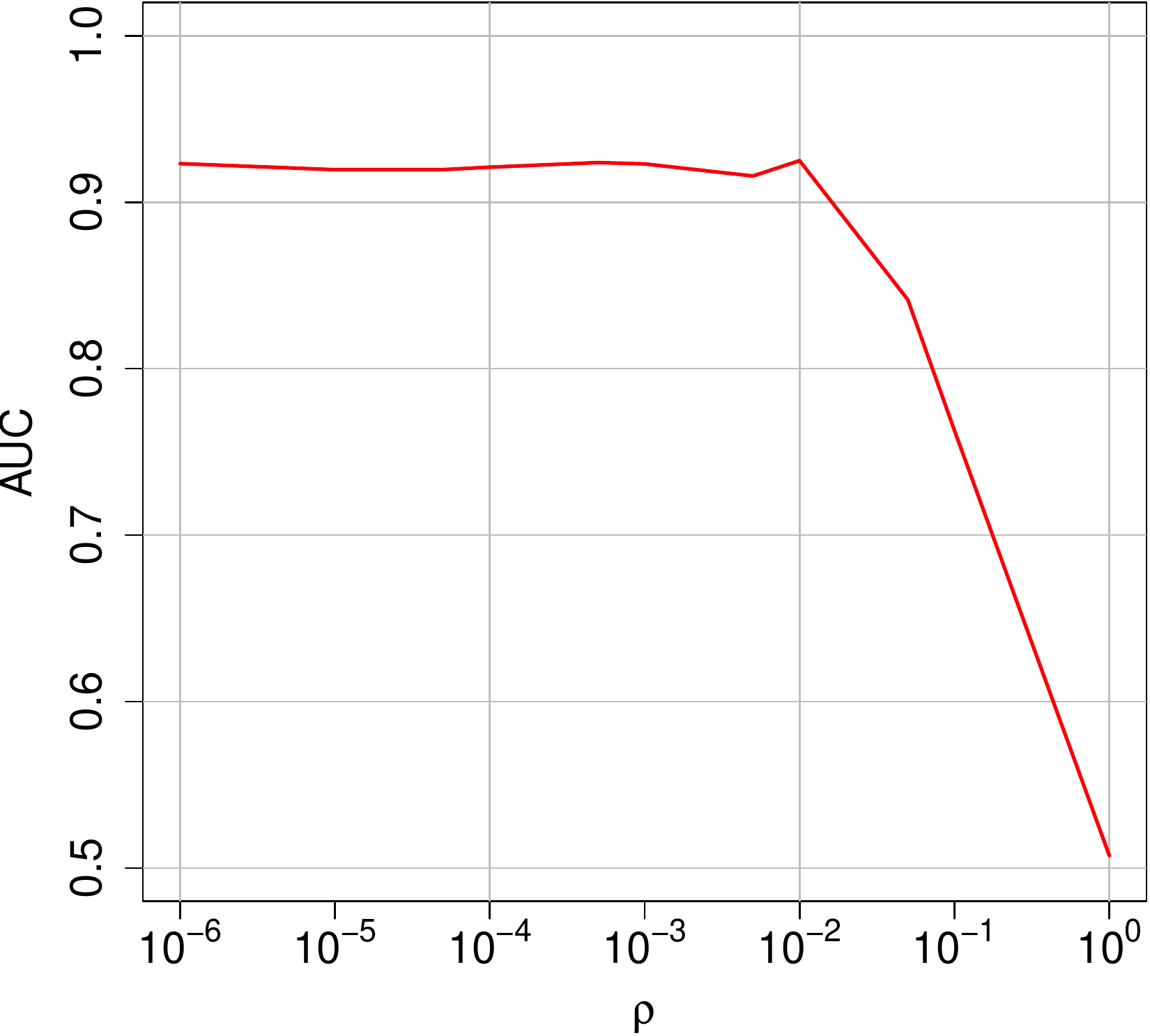}
       \label{fig:dbpedia-rho}
  }~
  \subfigure[Wikipedia-SVO]{
       \includegraphics[width=0.29\textwidth]{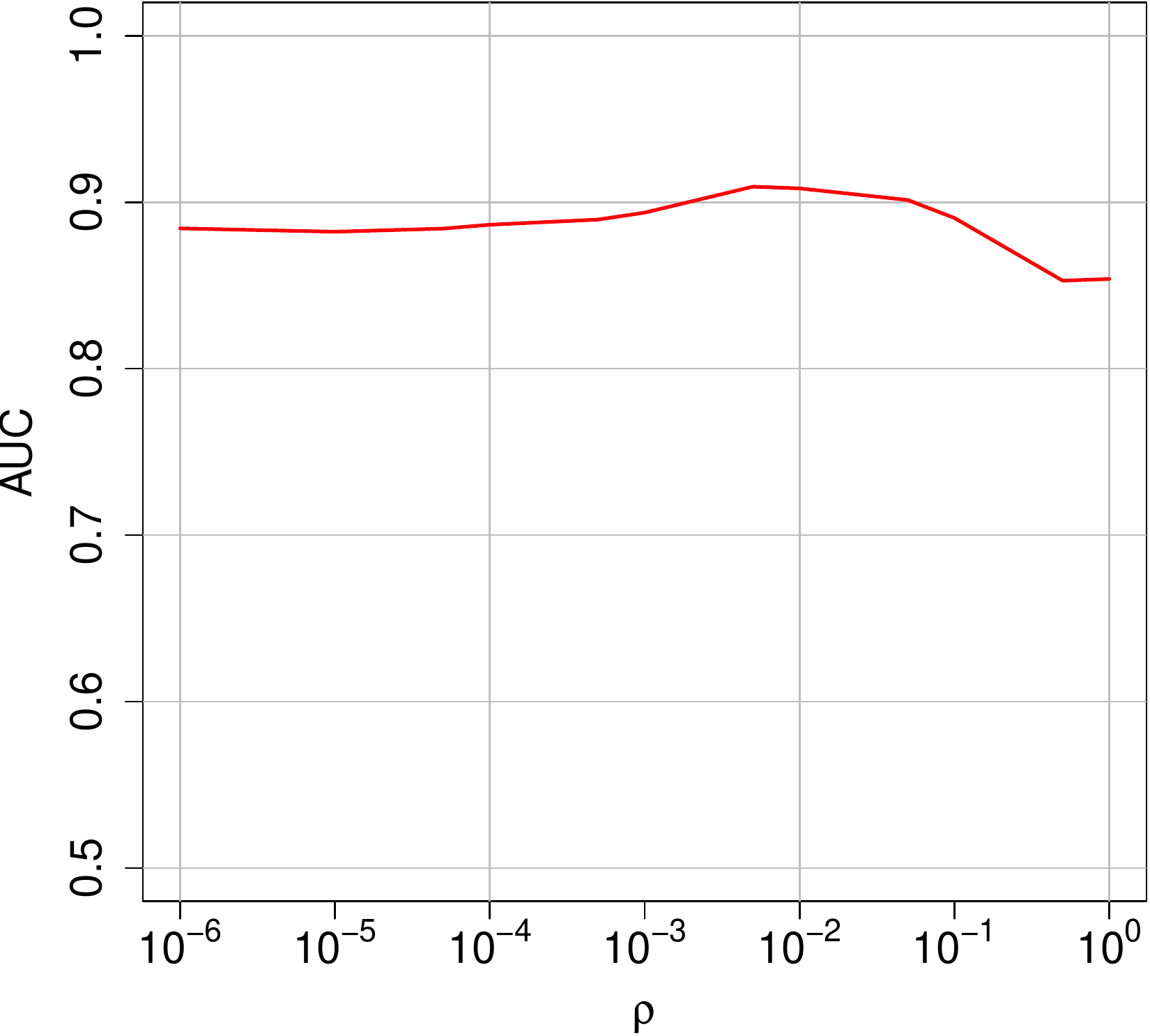}
       \label{fig:svo-rho}
  }~
  \subfigure[YAGO]{
       \includegraphics[width=0.29\textwidth]{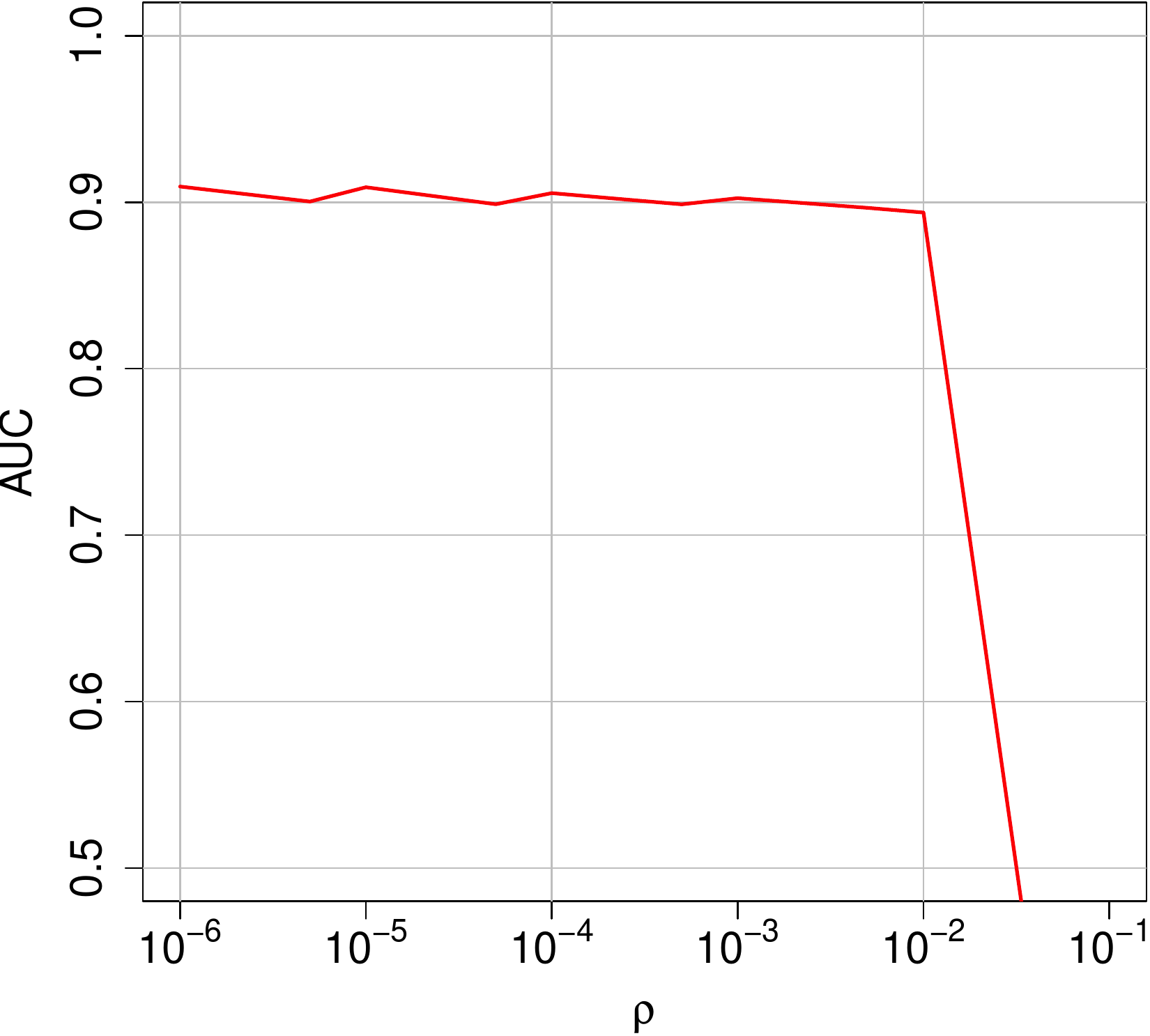}
       \label{fig:yago-rho}
  }
  \caption{Sensitivity to the $\rho$ hyperparameter.}
  \label{fig:sensitivity}
\end{figure*}

\noindent\textbf{Sensitivity to the $\rho$ hyperparameter.~} Our method ConsMRF introduces a new hyperparameter, $\rho$, which controls the extent to which target specific parameters $\A_r$ are regularized against the global entity latent features $\vect{Z}$. This means that higher values for $\rho$ tend to diminish the effect of target specific parameters since it forces each $\A_r$ to have similar values to $\vect{Z}$. Finally, note that $\rho$ acts as a step size for the $\vect{V}_r$ parameters (cf. Equation \ref{eq:updateV}), thus another side effect of large $\rho$ values is that it can lead to numerical problems or cause the algorithm to diverge. This explains the drop in performance seen on Figure \ref{fig:dbpedia-rho} and Figure \ref{fig:yago-rho}. The sensitivity of ConsMRF to the hyperparameter $\rho$ on the three datasets can be seen in Figure \ref{fig:sensitivity}. 

\noindent\textbf{Reproducibility of the experiments.~} All datasets used in our experiments are publicly available. A reference implementation for ConsMRF will be made available for download upon paper acceptance.

\section{Conclusion and Future Work}

Previous work has shown that multi-relational factorization models that optimize specifically for each target relation achieve better predictive performance. In this work, we have taken the idea of employing target specific parameters one step further by means of consensus optimization and the Alternating Direction Method of Multipliers (ADMM). Our novel method, ConsMRF, takes advantage of the predictive power of target specific parameters with a simple and efficient algorithm capable of scaling to large datasets. 

We have shown that ConsMRF can achieve state-of-the-art performance in much less time. In addition, to the best of our knowledge, ConsMRF the first principled method able to parallelize the learning of multi-relational factorization models. 

ConsMRF does not require a careful optimization of a large number of hyperparameters to balance the contribution of the different relations for the target prediction, which is a key advantage over previous approaches like DMF and CATSMF. 
 
Due to the partitioning of the problem across the relations, the work distribution between different threads might be unbalanced, especially if there is much more data about some relations than others. As future work we plan to achieve even better runtime performance improvements by exploiting different strategies for data partitioning in order to obtain a more balanced workload and thus even greater speedup from the parallelization. 

\bibliographystyle{IEEEtran}
\bibliography{IEEEabrv,main}
\flushend

\end{document}